\def\year{2018}\relax
\documentclass[letterpaper]{article} 
\usepackage{aaai18}  
\usepackage{times}  
\usepackage{helvet}  
\usepackage{courier}  
\usepackage{url}  
\usepackage{graphicx}  
\usepackage{amsmath}
\usepackage{amssymb}
\usepackage{booktabs}
\usepackage{makecell}
\usepackage{multirow}
\usepackage{tabularx}
\usepackage{color}
\newcolumntype{Y}{>{\centering\arraybackslash}X}

\def\refEqn#1{Eqn.~(\ref{#1})}
\def\refFig#1{Fig.~\ref{#1}}
\def\refTable#1{Table~\ref{#1}}
\DeclareMathOperator*{\argmax}{arg\,max}
\DeclareMathOperator*{\argmin}{arg\,min}

\begin{document}
	%
	\title{Dilated FCN for Multi-Agent 2D/3D Medical Image Registration}
	
 	\author{
 		Shun Miao$^{1*}$,
 		Sebastien Piat$^1$,
 		Peter Fischer$^2$, 
 		Ahmet Tuysuzoglu$^1$, 
 		Philip Mewes$^2$, 
 		\\ {\Large \bf
 			Tommaso Mansi$^1$,
 			Rui Liao$^1$}\\
 		$^1$Siemens Healthineers, Medical Imaging Technologies, Princeton, NJ, USA\\
 		$^2$Siemens Healthineers, Forchheim, Germany
 	}
	
	\newcommand\blfootnote[1]{%
		\begingroup
		\renewcommand\thefootnote{}\footnote{#1}%
		\addtocounter{footnote}{-1}%
		\endgroup
	}
	
	\maketitle
	\begin{abstract}
		2D/3D image registration to align a 3D volume and 2D X-ray images is a challenging problem due to its ill-posed nature and various artifacts presented in 2D X-ray images. In this paper, we propose a multi-agent system with an auto attention mechanism for robust and efficient 2D/3D image registration. Specifically, an individual agent is trained with dilated Fully Convolutional Network (FCN) to perform registration in a Markov Decision Process (MDP) by observing a local region, and the final action is then taken based on the proposals from multiple agents and weighted by their corresponding confidence levels. The contributions of this paper are threefold.	First, we formulate 2D/3D registration as a MDP with observations, actions, and rewards properly defined with respect to X-ray imaging systems. Second, to handle various artifacts in 2D X-ray images, multiple local agents are employed efficiently via FCN-based structures, and an auto attention mechanism is proposed to favor the proposals from regions with more reliable visual cues. Third, a dilated FCN-based training mechanism is proposed to significantly reduce the Degree of Freedom in the simulation of registration environment, and drastically improve training efficiency by an order of magnitude compared to standard CNN-based training method. We demonstrate that the proposed method achieves high robustness on both spine cone beam Computed Tomography data with a low signal-to-noise ratio and data from minimally invasive spine surgery where severe image artifacts and occlusions are presented due to metal screws and guide wires, outperforming other state-of-the-art methods (single agent-based and optimization-based) by a large margin.
	\end{abstract}
	
	\section{Introduction}
	
	
	The goal of 2D/3D medical image registration is to find the 6 Degree of Freedom (DoF) pose of a 3D volume (e.g. Computed Tomography (CT), Mangnetic Resonance Imaging (MRI) etc), to align its projections with given 2D X-ray images. Reliable 2D/3D registration is a key enabler for image-guided surgeries in modern operating rooms. It brings measurement and plannings done on the pre-operative data into the operating room, and fuse it with intra-operative live 2D X-ray images. It can be used to provide augmented reality image guidance for the surgery, or provide navigation for robotic surgery. Despite that 2D/3D registration has been actively researched for decades~\cite{markelj2012review}, it remains a very challenging and unsolved problem, especially in the complex environment of hybrid operating rooms. 
	\blfootnote{\textbf{Disclaimer}: This feature is based on research, and is not commercially available. Due to regulatory reasons its future availability cannot be guaranteed.} 
	
	\begin{figure*}
		\centering
		\includegraphics[width=\linewidth]{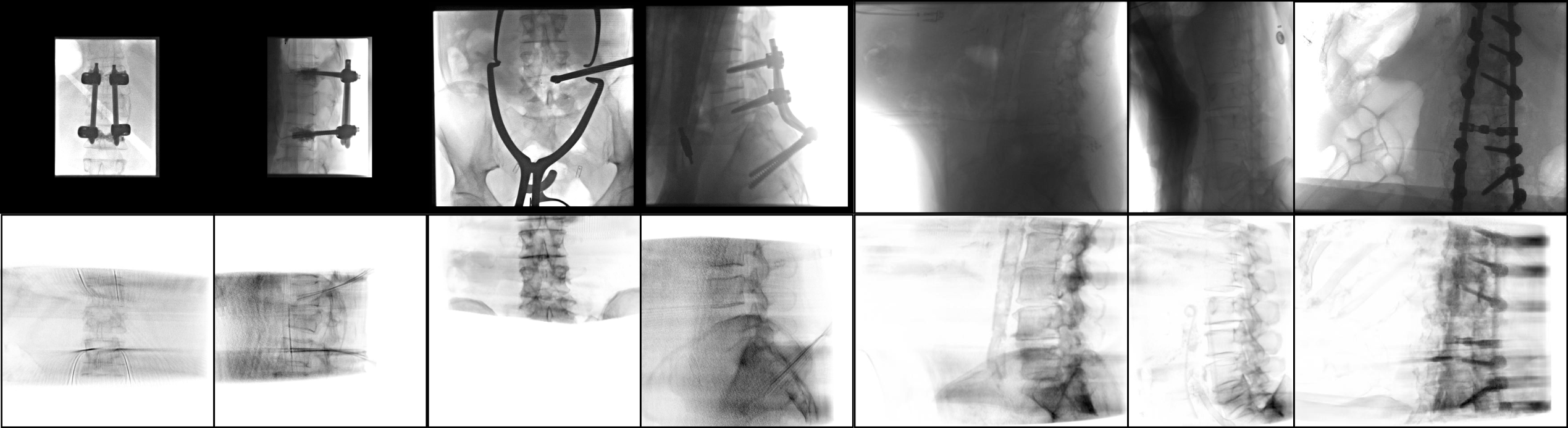}
		\caption{Example X-ray images (top row) and corresponding Digitally Reconstructed Radiographs (DRRs) (bottom row). The first four X-ray images are clinical data from spine surgery, which contain various highly opaque metal objects and have very different Field-of-Views (FoVs). The last three X-ray images are from CBCT data, which have a relatively low SNR due to a very small dynamic range.}
		\label{fig:example_images}
	\end{figure*}
	
	A Markov Decision Process (MDP) formulation for 3D/3D image registration was recently introduced and resulted in a significant robustness improvement~\cite{liao2017artificial}. The original formulation however has three major limitations that make it ineffective for 2D/3D registration in a clinical setup. First, it requires the rough location and size of the target object to be registered as a prior, in order to extract a local region around the target as the agent's observation. However, in 2D/3D registration, the location and size of the target object can vary significantly in 2D X-ray images, due to variations in C-arm geometry and imaging protocols such as collimation factors. Second, there could be various artifacts / interference coming from medical devices in 2D X-ray images and simulation of all of them in the training samples is not practical. Therefore an auto attention mechanism is needed in order to be able to inherently detect the regions with more reliable visual cues to drive registration, and this mechanism is not provided in ~\cite{liao2017artificial}. Third, training data need to be sampled extensively from the registration environment with a high DoF (i.e., environmental DoFs include location of agent's observation and pose of the 3D volume), which is associated with an high computational cost. In fact, five million samples are needed as reported in~\cite{liao2017artificial}, even after using location prior knowledge to reduce the DoF by 3. Since data sampling grows exponentially with the DoF, without the location prior knowledge, computational cost would be prohibitively high for our 2D/3D registration problem. 
	
	In this paper, we propose a multi-agent system with an auto attention mechanism to address the above three limitations of agent-based registration, and apply it to a challenging 2D/3D registration application for minimally invasive spine surgery. In particular, multiple agents are employed to observe multiple regions of the image, and the system adaptively favors proposals from regions with more distinct visual cues for registration. We furthermore propose a policy network architecture with separate encoding of the fixed and moving images, and a dilated Fully Convolutional Network (FCN) based training strategy to train all observation regions in each back propagation. This strategy significantly reduces the DoFs of the registration environment, and as a result training efficiency is improved by an order of magnitude compared to standard CNN-based training method. The proposed FCN structure also naturally supports our multi-agent system in the application phase for efficient 2D/3D registration. We demonstrate that when there are severe image artifacts and occlusions, e.g. due to metal screws and guide wires, the proposed method significantly outperforms single agent-based method and other state-of-the-art optimization-based methods by reducing the gross failure rate (GRF) by 10 fold.  
	
	\section{Related Work}
	
	
	\subsection{Optimization-based 2D/3D Registration}
	
	The most widely adopted formulation for 2D/3D registration is to solve it as an optimization problem to minimize a cost function that quantifies the quality of alignment between the images being registered~\cite{markelj2012review}\cite{gendrin2012monitoring}\cite{schmid2014segmentation}\cite{miao2011hybrid}. The success of these approaches depends on the global optimization of the cost function. Existing image-based cost functions are mostly based on low level metrics, e.g., Mutual Information (MI)~\cite{maes1997multimodality}, Cross Correlation (CC)~\cite{knaan2003effective}, Gradient Correlation (GC)~\cite{de20163d} and etc. These metrics compare images directly at the pixel level without understanding the higher level structures in the image.
	As a result, on images with a low signal-to-noise ratio (SNR) and/or severe image artifacts like examples shown in \refFig{fig:example_images}, they often have numerous local minima, which makes it extremely challenging to locate the global solution using mathematical optimization strategy, e.g. Powell's method, Nelder-Mead, BFGS, CMA-ES. 
	Although there are global optimization methods, e.g., simulated annealing \cite{van1987simulated} and genetic algorithm \cite{deb2002fast}, they require comprehensive sampling of the parameter space, which leads to a prohibitively high computational cost. Few attempts have been made to seek heuristic semi-global optimization to seek a proper balance between robustness and computational cost~\cite{uneri2017intraoperative}\cite{lau2006global}.
	
	
	\subsection{Learning-based Image Registration}
	
	Several attempts have been made recently to tackle 2D/3D registration problems using learning-based approaches. Hand crafted features and simple regression modules like Multi-Layer Perceptron and linear regression were proposed to regress 2D/3D registration parameters~\cite{gouveia2015registration}~\cite{chou20132d}. A CNN-based regression approach was introduced in~\cite{miao2016cnn} to solve 2D/3D registration for 6 DoF device pose estimation from X-ray images, and later on a domain adaptation scheme was proposed ~\cite{zheng2017learning} to improve the performance generalization of the CNN model on unseen real data. In~\cite{wohlhart2015learning}, a range image descriptor was learned via CNN to differentiate 3D pose of the object, and was used for 6 DoF pose estimation. {While the CNN-based methods improve robustness of 2D/3D registration, they are limited to registering only highly opaque objects with a fixed shape described by a CAD model (e.g., rigid metal devices), because they aimed at explicitly modeling the appearance-pose relationship. Therefore, it cannot solve 2D/3D registration of anatomical structures with varying shapes across patients.}
	
	Learning-based approaches have also been explored for 2D/2D and 3D/3D image registration, where the goal is to find the spatial transformation to align images of the same dimension. Unsupervised learning using CNN was proposed in \cite {wu2006learning} to extract features for deformable registration. These features however were extracted separately from the image pairs and therefore may not be optimal for registration purpose. Optical flow estimation between 2D RGB images has been proposed using CNN via supervised learning in \cite{weinzaepfel2013deepflow}\cite{dosovitskiy2015flownet}\cite{ilg2016flownet}. A learnable module, called spatial transformer network (STN), was introduced in \cite{jaderberg2015spatial}. The focus of STN was not an accurate alignment of two images, but a rough transformation of a single input image to a canonical form. In a recent work, rigid-body 3D/3D image registration is formulated as a MDP and a policy network is trained to perform image registration~\cite{liao2017artificial}. Motivated by this work, we design a 2D/3D registration system by utilizing multiple agents coupled with an auto attention mechanism for high robustness, and train the agent using dilated FCN-based strategy for high efficiency.
	
	
	\section{Background}
	
	\subsection{2D/3D Registration}
	
	Given a 3D CT volume $J: \mathbb{R}^3 \to \mathbb{R}$, a projection image can be calculated following the X-ray imaging model~\cite{bushberg2011essential}:
	\begin{equation}
		H_T(\boldsymbol{p}) = \int{J \Big( T^{-1} \circ {\boldsymbol{L}(\boldsymbol{p},r)} \Big) dr},
		\label{eqn:xray3}
	\end{equation}
	where $I(\boldsymbol{p})$ is the intensity of the synthetic X-ray image at point $\boldsymbol{p}$, ${\boldsymbol{L}(\boldsymbol{p},r)}$ is the line connecting the X-ray source and the point $\boldsymbol{p}$, parameterized by $r$, and $T: \mathbb{R}^3 \to \mathbb{R}^3$ is the geometric transformation of the 3D volume. Such projection image is referred to as DRR, and can be computed using the Ray-Casting algorithm~\cite{kruger2003acceleration}.
	
	In 2D/3D registration problems, a 3D volume $J(\cdot)$, a 2D X-ray image $I(\cdot)$ and the camera model of the X-ray images $\boldsymbol{L}(\cdot)$ are given. The goal is to find the transformation $T$ that aligns the projection of the 3D volume $H_T(\cdot)$ with the X-ray image $I(\cdot)$. Due to the ambiguity in matching a 3D volume with a single projected 2D image, multiple X-ray images from different projection angles are often employed in 2D/3D registration. In such cases, the goal is to find the transformation $T$ that aligns all DRR and X-ray image pairs, denoted as $H_{i,T}(\cdot)$ and $I_i$, where $i$ denotes the index of the X-ray image.
	
	\subsection{Special Euclidean Group SE(3)}
	
	Special Euclidean group SE(3) is the set of $4\times4$ matrices corresponding to translations and rotations. The tangent space of SE(3) is described using the Lie algebra se(3), which has six generators corresponding to the derivatives of translation and rotation along/around each of the standard axes. An element of se(3) is then represented by multiples of the generators
	\begin{align}
		\boldsymbol{\delta} & = (\boldsymbol{u}, \boldsymbol{v}) \in \mathbb{R}^6 
		\label{eqn:se3_params} \\
		\boldsymbol{\delta}_{\times} & = u_1 G_1 + u_2 G_2 + u_3 G_3 +  \nonumber \\
		& v_1 G_4 + v_2 G_5 + v_3 G_6 \in \text{se}(3),
	\end{align}
	where $(G_1,G_2,G_3)$ are the translation generators, and $(G_4, G_5, G_6)$ are the rotation generators. Matrix exponential and logarithm can be taken to convert elements between SE(3) and se(3).
	
	\section{Proposed Method}
	
	\begin{figure}
		\centering
		\includegraphics[width=\linewidth]{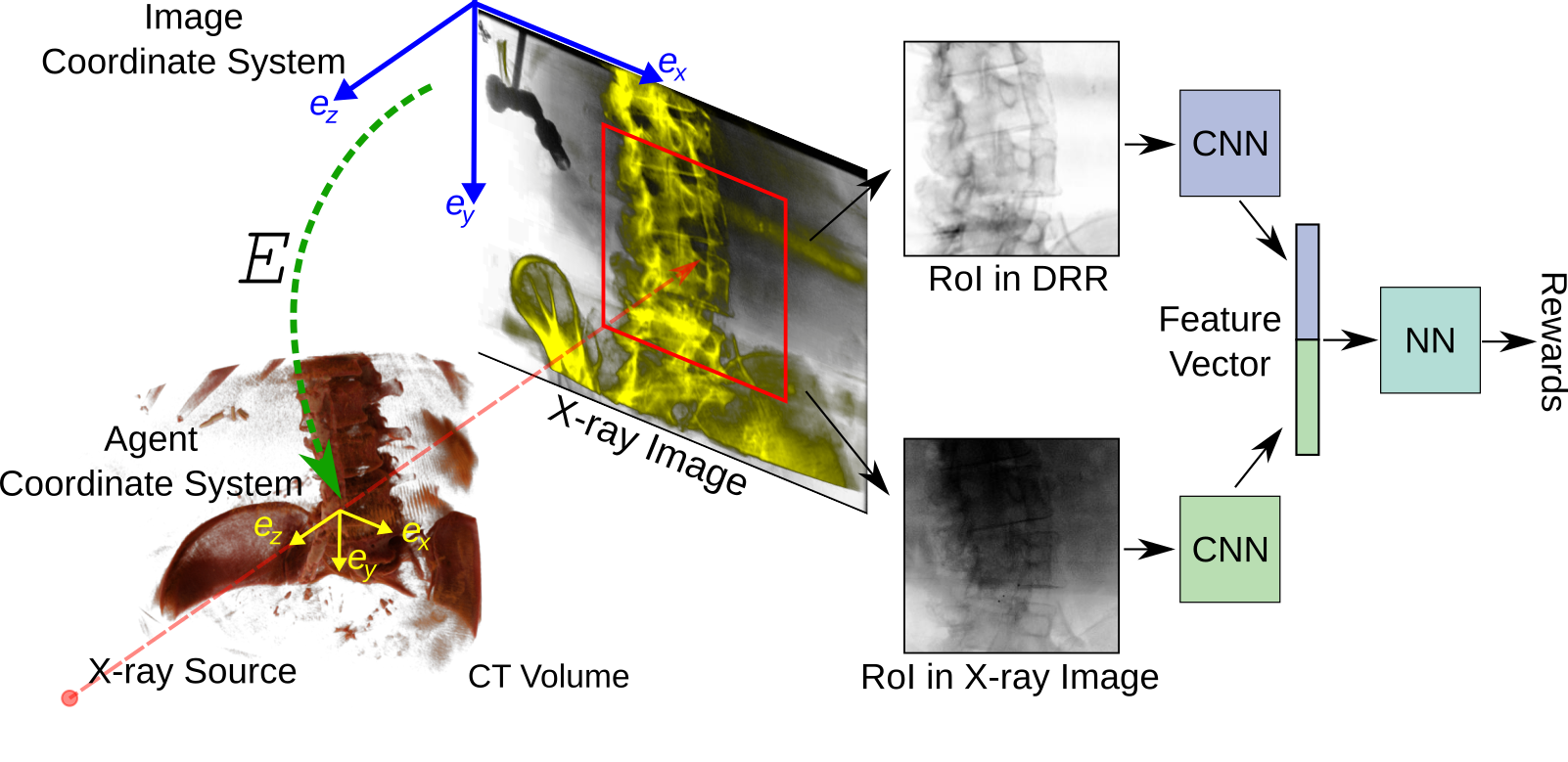}
		\caption{Policy network with encoder CNN and decoder neural network (NN). An ROI centered on the projection of the agent's location is extracted from the X-ray image and DRR. The extracted regions are encoded using CNN to obtain feature vectors, which are concatenated and decoded by a NN to obtain reward estimation.}
		\label{fig:system_overview}
	\end{figure}
	
	
	\subsection{Markov Decision Process}
	
	We cast the problem of finding $T$ to align the DRR $H_T(\cdot)$ with a X-ray image $I(\cdot)$ as a MDP, which is defined by a 5-tuple $\{\mathcal{T}, \mathcal{A}, P_{\cdot}(\cdot), R_{\cdot}(\cdot), \gamma\}$, where $\mathcal{T}$ is the set of possible states (i.e., transformations in SE(3)), $\mathcal{A}$ is the set of actions (i.e., modification of the transformation), $P_A(T)$ is the state obtained by taking action $A$ in state $T$, $R_A(T)$ is the reward received by taking action $A$ in state $T$, and $\gamma$ is the discount factor that controls the importance of future rewards. 
	With the action space $\mathcal{A}$ and the reward scheme $R_\cdot(\cdot)$ defined (details are provided in the next sections), the core problem of MDP is to find a policy $\pi(\cdot)$ that specifies the optimal action $\pi(T_t)$ to be taken at state $T_t$ to maximize the long term reward:
	\begin{equation}
		\sum_{t=0}^{\inf} {\gamma^t R_{A_t}(T_t)}, \quad \text{where we choose }A_t=\pi(T_t).
	\end{equation}
	
	\subsection{Action Space}
	
	To make the policy learnable, we define the action space based on the X-ray projection geometry such that each action is correlated with a specific appearance change of the DRR that is largely independent of C-arm geometry. Specifically, the transformation $T$ is described in the image coordinate system,  which has its origin at the upper left corner of the image, its $(x,y)$ axes along the image edges, and its $z$ axes perpendicular to the image (illustrated in \refFig{fig:system_overview}). We further define an agent coordinate system with the same orientation as the image coordinate system and an origin that can be selected individually (details on the selection of the origin will be discussed later). The coordinate transform from the image coordinate system to the agent coordinate system is denoted as $E$.
	The transformation $T_t$ is then describe in the agent coordinate system, written as $E \circ T_t$, and the action space is defined as small movements in the tangent space of SE(3) at $E \circ T_t$, parameterized by se(3).
	Specifically, the action space contains 12 actions of positive and negative movements along the 6 generators of se(3):
	\begin{equation}
		\mathcal{A} = \{-\lambda_1 G_1, \lambda_1 G_1, \dots, -\lambda_6 G_6, \lambda_6 G_6\},
	\end{equation}
	where $\lambda_i$ is the step size for the action along the generator $G_i$. Application of an action $A \in \mathcal{A}$ is represented as
	\begin{equation}
		T_{t+1}=E^{-1} \circ \exp(A) \circ E \circ T_t.
	\end{equation}
	Since the actions need relatively small step size in order to achieve high accuracy, we set $\lambda_{1,2,3}$ to be 1 to get a step size of 1 mm in translation, and $\lambda_{4,5,6}$ to be $\pi / 180=0.0174$ to get a step size of 1 degree in rotation. Our definition of the action space ensures that the translation actions causes 2D shift and zooming of the DRR, and rotation actions causes rotation of object in the DRR around the agent's origin. Therefore the image appearance change of the DRR for a given action is largely independent of the underlying C-arm geometry. Note there is no action for terminating the MDP. Instead, we run the agent for a fixed number of steps (i.e., 50 in our experiments).
	
	\subsection{Reward System}
	
	In standard MPD, the optimization target is a long term reward, i.e., an accumulation of discounted future reward, due to the difficulty of forging a reward system that directly associate the immediate reward with the long term goal. For 2D/3D registration, however, we can define a distance-based reward system such that the immediate reward is tied with the improvement of the registration. The reward scheme is defined as the reduction of distance to the ground truth transformation in the agent coordinate system:
	\begin{equation}
		R_A(T) = D(E\circ T, E\circ T_g) - D(E\circ T', E\circ T_g),
		\label{eqn:reward}
	\end{equation}
	where $T$ and $T'$ are transformations before and after the action, $T_g$ is the ground truth transformation. The distance metric $D(\cdot, \cdot)$ is defined as the geodesic distance of two transformations on SE(3) \cite{hartley2011l1}:
	\begin{equation}
		\begin{split}
			D(T_1, T_2) &= \| \log (T_2 \circ T_1^{-1}) \|_F \\
			            &= \left( 2\| \boldsymbol{u} \|_2^2 + \| \boldsymbol{v} \|_2^2 \right) ^ {\frac{1}{2}},
		\end{split}
		\label{eqn:tdist}
	\end{equation}
	where $\log(\cdot)$ takes $T_2 \circ T_1^{-1} \in \text{SE}(3)$ into se(3), $\boldsymbol{u}$ and $\boldsymbol{v}$ are rotation and translation coefficients of $\log (T_2 \circ T_1^{-1})$ as described in \refEqn{eqn:se3_params}. Because the units for rotation and translation are radian and mm, the distance impact of rotation is too small. Therefore, we scale the rotation coefficients $\boldsymbol{v}$ by $180/\pi$ to balance the impacts of rotation and translation.
	
	Since the distance $D(T, T_g)$ measures the distance to the ground truth, a greedy policy that maximizes the immediate reward (i.e., minimizes the distance to the ground truth) would lead to the correct registration.
	\begin{equation}
		\pi(T) = \argmax_A R_{A}(T).
		\label{eqn:policy}
	\end{equation}
	It can also be seem as a special case of MPD with the discount factor $\gamma = 0$.
	
	\subsection{Policy Learning via Dilated FCN}
		
	\begin{figure*}
		\centering
		\includegraphics[width=0.78\linewidth]{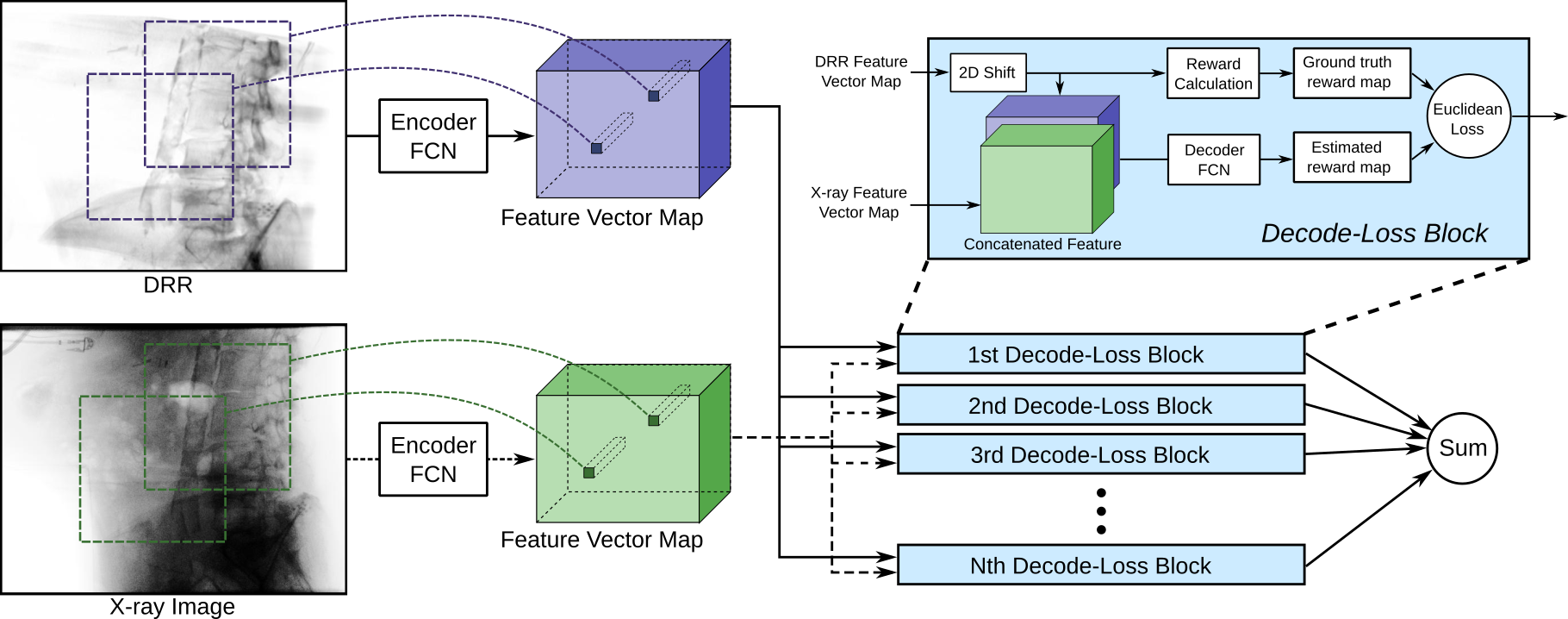}
		\caption{Training of the reward network using dilated FCN. The encoder CNN and decoder NN are converted to equivalent dilated FCNs. Densely overlapping ROIs are encoded to a dense feature vector map. The feature vector map of DRR is then randomly shifted N (N=4 in our experiments) times to simulate translations of the 3D volume in the imaging plane. A ground truth dense reward map is calculated from each 2D shift combined with the 3D transformation. Euclidean loss comparing the estimated and ground truth rewards is used for training.}
		\label{fig:fcn_system}
	\end{figure*}
	
	Since we seek a greedy policy that maximizes the immediate reward, we model the reward function as a deep neural network, as shown in \refFig{fig:system_overview}, which can be learned via supervised learning.
	The input of the network is an observation of the current state, which consists of an observed region of fixed size (i.e. $61\times 61$ pixels with 1.5$\times$1.5 mm pixel spacing in our experiment) from both X-ray and DRR, referred to as Region of Interest (ROI). The ROI of an agent is centered on the projection of its origin, so that rotation actions cause rotations with respect to the center of the ROI.
	Such ROI center ensures that the same action always cause the same visual change in the observation, leading to a consistent mapping between the observation and rewards of different actions that can be modeled by a neural network. 
	We propose a network architecture that encodes X-ray and DRR separately and decodes them jointly to obtain the estimated reward, as shown in \refFig{fig:system_overview} and \refTable{tab:netconfig}.
	
	\setlength{\tabcolsep}{7pt}
	\begin{table}[t]
		\centering
		\caption{Layer configurations for encoder/decoder CNNs and their equivalent dilated FCNs. Parameters for convolutional layers are written as $m \times n \times f$, where $n \times m$ is the convolution kernal size, $f$ is the number of feature maps. s$k$ indicates that the layer has input stride $k$, and d$k$ indicates that the filter kernel is dilated $k$ times. All convolutional layers have zero padding. SELU activation function is applied after all layers except for the input and output layers. The column "output size" specifies the output sizes for CNN.}
		\label{tab:netconfig}
		\begin{tabular}{lllll}
			\toprule
			& \makecell{Layer \\ Name} & \makecell{Output \\ Size} & \makecell{Original \\ CNN} & \makecell{Equivalent \\ FCN}  \\       \midrule
			\multirow{11}{*}{\rotatebox[origin=c]{90}{Encoder}} 
			&input      & 61x61       & -           & -           \\ \cline{2-5}
			&conv1      & 59x59       & 3x3x32      & 3x3x32,     \\ \cline{2-5}
			&conv2      & 57x57       & 3x3x32      & 3x3x32,     \\ \cline{2-5}
			&conv3      & 27x27       & 3x3x64,s2   & 3x3x64,d2   \\ \cline{2-5}
			&conv4      & 25x25       & 3x3x64      & 3x3x64,d2   \\ \cline{2-5}
			&conv5      & 11x11       & 3x3x128,s2  & 3x3x128,d4  \\ \cline{2-5}
			&conv6      & 9x9         & 3x3x128     & 3x3x128,d4  \\ \cline{2-5}
			&conv7      & 3x3         & 3x3x256,s2  & 3x3x256,d8  \\ \cline{2-5}
			&fc1        & 1x1         & 1024        & 3x3x1024,d8 \\ \cline{2-5}
			&fc2        & 1x1         & 1024        & 1x1x1024,d8 \\ \cline{2-5}
			&output     & 1x1         & 128         & 1x1x128,d8  \\ \midrule
			\multirow{4}{*}{\rotatebox[origin=c]{90}{Decoder}}
			&input      & 1x1         & -           & -           \\ \cline{2-5}
			&fc1        & 1x1         & 1024        & 3x3x1024,d8 \\ \cline{2-5}
			&fc2        & 1x1         & 1024        & 1x1x1024,d8 \\ \cline{2-5}
			&output     & 1x1         & 6           & 1x1x6,d8    \\ \bottomrule
		\end{tabular}
	\end{table}
	
	Such CNN can be trained by supervised learning on data randomly sampled from the registration environment, similar to \cite{liao2017artificial}. In particular, given one pair of 2D and 3D data, training data of the CNN can be generated with 8 DoFs, including 2 DoFs for $(x,y)$ of the agent's origin ($z$ is set to be the mid-point of the 3D volume) corresponding to the location of the agent, and 6 DoFs for the transformation $T$ corresponding to the pose of the 3D volume. The ground truth rewards can be calculated following \refEqn{eqn:reward}. To make the learned reward function generalizable to unseen data, all 8 DoFs need to be sufficiently sampled during training, which has a prohibitively high computational cost. In \cite{liao2017artificial}, five million samples are needed to cover training data of 6 DoFs, and the sampling requirement grows exponentially with the DoF.
	
	We propose a dilated FCN-based training mechanism to reduce the DoF of training data from 8 to 4. First, the encoder and decoder CNNs are converted to equivalent dilated FCNs. The filter dilation technique introduced in~\cite{long2015fully} is used to make each pixel in the FCN output exactly the same as that produced using the corresponding CNN on its receptive field in the input images. Layer to layer correspondence of CNN and dilated FCN is shown in \refTable{tab:netconfig}. Training using dilated FCN is illustrated in \refFig{fig:fcn_system}. Since the encoder FCN encodes all ROIs in the input image, it can replace sampling $(x,y)$ of the agent's origin. The $(x,y)$ translation part of $T$ creates 2D shift of the DRR, which can be replaced by shifting the encoded feature map of the DRR during training. Therefore, 4 DoFs of training data can be avoided by using FCN-based training, reducing the number of total DoFs from 8 to 4. Since the complexity of the environment grows exponentially with its DoF, reducing the DoF by 4 can significantly improve the training efficiency.
	
	In the FCN-based training, the ground truth reward map needs to be calculated during training, following \refEqn{eqn:reward} and \refEqn{eqn:tdist}. Given a X-ray/DRR pair with known $T_g$ and $T$, the agent coordinate system $\{E_i\}$ is calculated for all ROIs within the input image by tracing the pixel back to the 3D volume. For each ROI, $(\boldsymbol{u}_i, \boldsymbol{v}_i)$ of $\log \left(E_i \circ T \circ (E_i \circ T_g)^{-1} \right)$ is pre-calculated. The effect of 2D shift of the DRR feature map is essentially adding the translation to $T$, which is equivalent to updating $\boldsymbol{u}_i \leftarrow \boldsymbol{u}_i + \Delta \boldsymbol{u}$, where $\Delta \boldsymbol{u}$ is the 2D shift represented in mm. The distance before action is then calculated using $(\boldsymbol{u}_i, \boldsymbol{v}_i)$ following \refEqn{eqn:tdist} for each ROI. Similarly, the distance after action is calculated following the same steps by replacing $T$ with $T' = E_i^{-1} \circ \exp(A) \circ E_i \circ T$. Note that the expensive computation of matrix logarithms can be pre-calculated for each X-ray/DRR pair, as it does not depend on the 2D shift.
	
	\subsection{Multi-agent System}
	\begin{figure}
		\centering
		\includegraphics[width=\linewidth]{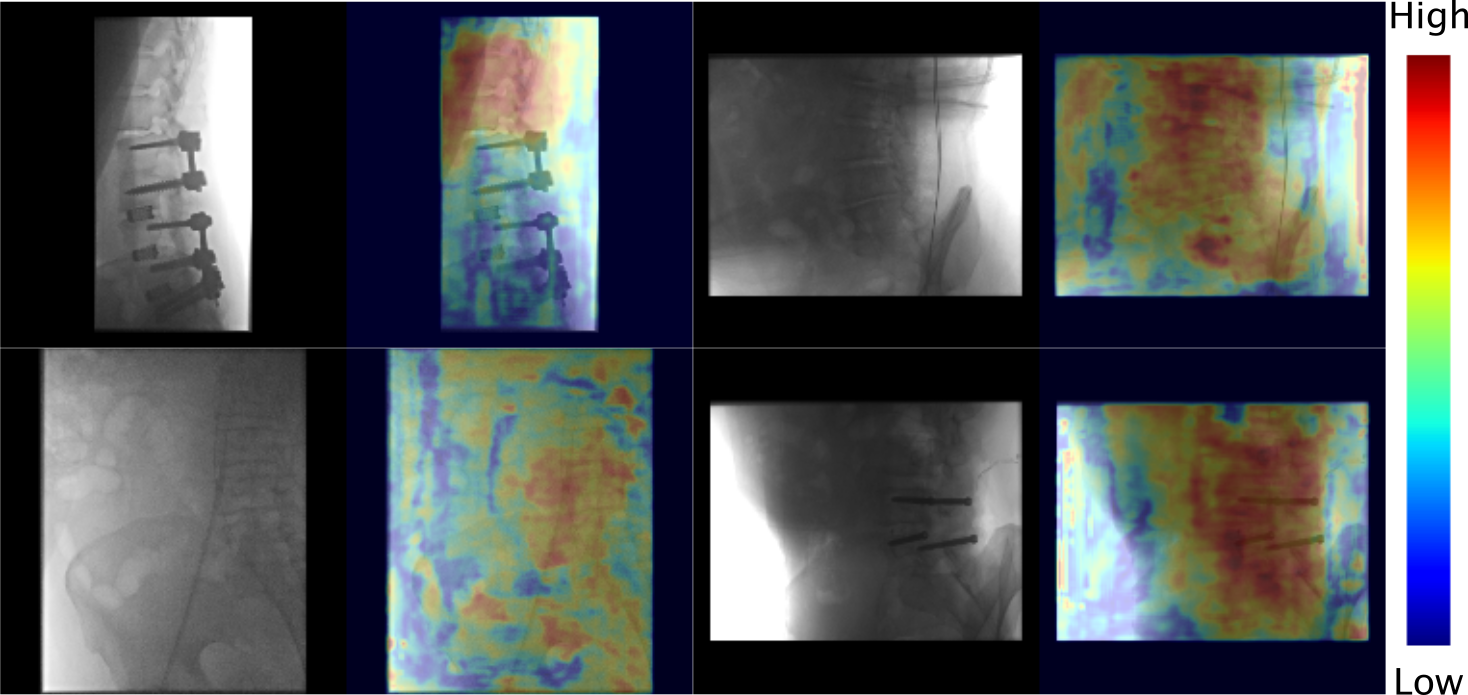}
		\caption{Confidence map of densely overlapping ROIs within the image. Color of each pixel indicates the confidence value from the ROI centered on this pixel.}
		\label{fig:conf_map}
	\end{figure}
	
	Since X-ray images during surgery and interventions can have very different Field-of-View (FoV) and contain many structures that do not match with the 3D image (e.g., medical devices), there can be many ROIs that do not contain reliable visual cues for registration (as shown in \refFig{fig:example_images}). Therefore we propose a multi-agent system to provide auto attention and adaptively choose most reliable ROIs during registration. Specifically, the FCN policy network is applied on the X-ray and DRR to produce a dense reward map, which contains estimated rewards for agents with all possible ROIs from the input images, denoted as $R_{i}(A)$, where $i$ is the index of the agent and $A \in \mathcal{A}$ is the action. For every agent, the maximum reward is calculated and the action associated with it is selected:
	\begin{equation}
		\begin{split}
			\hat{R}_i &= \max_{A\in \mathcal{A}}R_i(A), \\
			A_i &= \argmax_{A\in \mathcal{A}}R_i(A).
		\end{split}
	\end{equation}
	Since $\hat{R}_i$ is the anticipated reward of its selected action $A_i$, it represents the agent's confidence on the action. \refFig{fig:conf_map} shows a strong correlation between the confidence score $\hat{R}_i$ and the quality of the corresponding ROI: for image with a large FoV, the confidence score is high on spine (i.e., good visual cue for registration) and low on soft tissue; when severe occlusion is presented due to medical devices, the occluded area has low confidence scores.
	
	\begin{figure}
		\centering
		\includegraphics[clip, trim=1mm 4mm 1mm 4mm, width=0.9\linewidth]{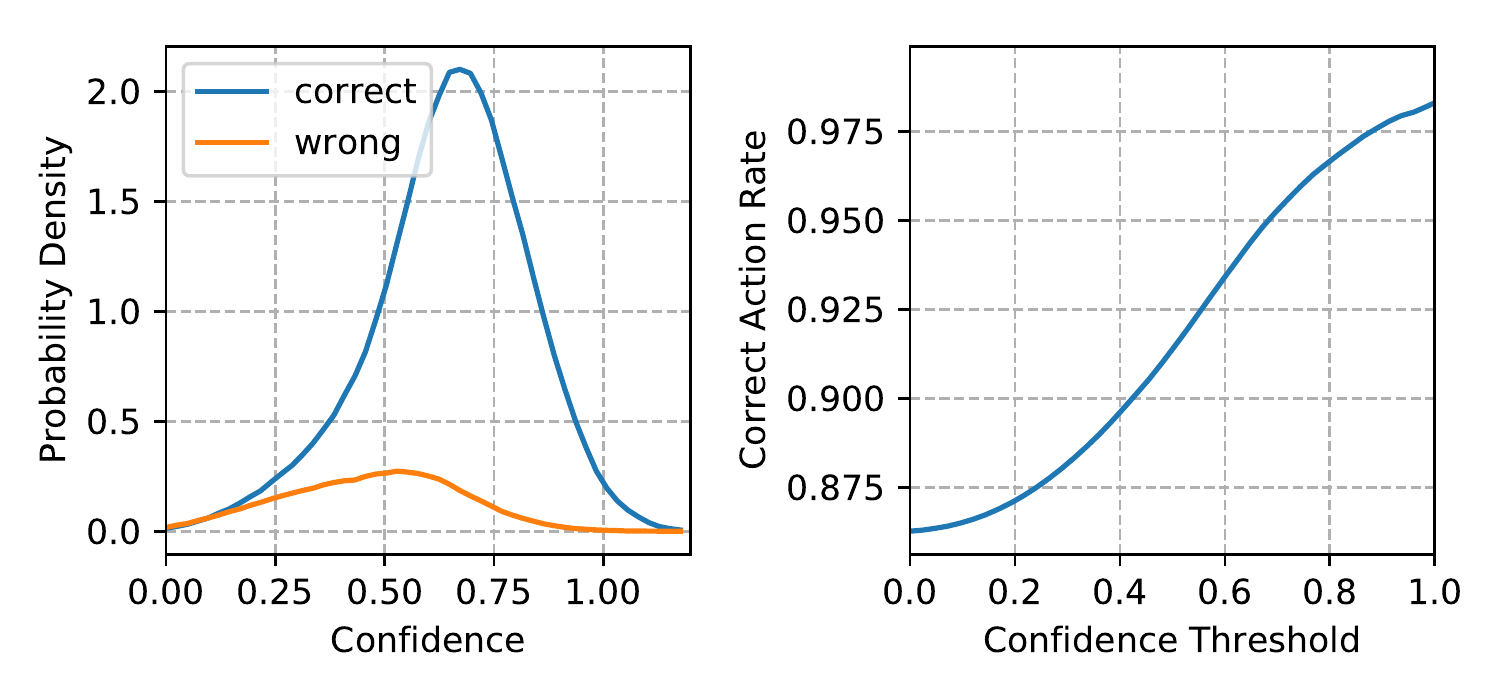}
		\caption{\textbf{Left}: Confidence distribution of agents producing correct and wrong actions. \textbf{Right}: Relation between the percentage of correct action among selected action and the confidence threshold used for selecting action.}
		\label{fig:conf_dist}
	\end{figure}
	
	We propose to use the confidence score $\hat{R}_i$ to derive an attention mechanism that only selects agents with $\hat{R}_i$ above a certain confidence threshold. To determine the threshold, the behavior of the confidence score was analyzed on validation data. Actions are categorized into correct and wrong, based on their impact on the registration (i.e., increase or decrease the distance to ground truth). \refFig{fig:conf_dist} shows the distribution of the confidence scores from agents producing correct and wrong actions, and relation between the percentage of correction action among selected actions and the confidence threshold used for selecting actions. Based on the analysis, we selected a confidence threshold (i.e., 0.67 in our experiment) such that the correct rate of selected actions is above 95\%. To avoid the scenario that too few agents are selected for a given test image, if less than 10\% of the agents have a confidence score above this threshold, the top 10\% agents will then be selected. After the agents are selected, denoted as $\mathcal{I}$, actions from the selected agents are further aggregated by L2 chordal mean to obtain the final actions:
	\begin{equation}
		\hat{A} = \argmin_{A \in SE(3)} \sum_{i \in \mathcal{I}} \| A_i - A \|_F^2.
		\label{eqn:aggregation}
	\end{equation}
	The L2 chordal mean can be solved globally in close form as described in \cite{hartley2011l1}.

	\section{Experiment and Result}
	
	We applied the proposed method on a clinical application of 2D/3D registration during minimally invasive spine surgery, which aims at registering spine in 3D cone beam CT (CBCT) and two X-ray images acquired from different angles. This is a challenging problem because surgical objects like screws and guide wires can be presented separately in the 3D and 2D images, creating severe image artifacts and occlusion of the target object (examples are shown in \refFig{fig:example_images}). 
	
	\subsection{Training}
	
	During minimally invasive spine surgery, the initial pose offset between the CBCT and the X-ray images can be up to 20 mm in translation and 10 degrees in rotation. Therefore, we train the agents to perform registration starting from within this range. In particular, the X-ray/DRR pairs used for training have random rotation offset up to $10^\circ$, and the DRR feature map is randomly shifted up to 20 mm during training. The training data was generated from 77 CBCT data sets, where each CBCT data set consists of one CBCT and $\sim$350 X-ray images used for reconstructing the CBCT. From each CBCT data set, we extracted 350 X-ray/DRR pairs. Since the number of CBCTs is limited, we also generated pairs of synthetic X-ray image and DRR from 160 CTs as additional training data, where 200 pairs where generated from each CT. In total, our training data consist of 58,950 data, i.e., 26,950 CBCT data and 32,000 synthetic data. Training was performed on a Nvidia Titan Xp GPU using pyTorch.
	
\begin{figure}
		\centering
		\includegraphics[clip, trim=1mm 2mm 1mm 2mm, width=0.9\linewidth]{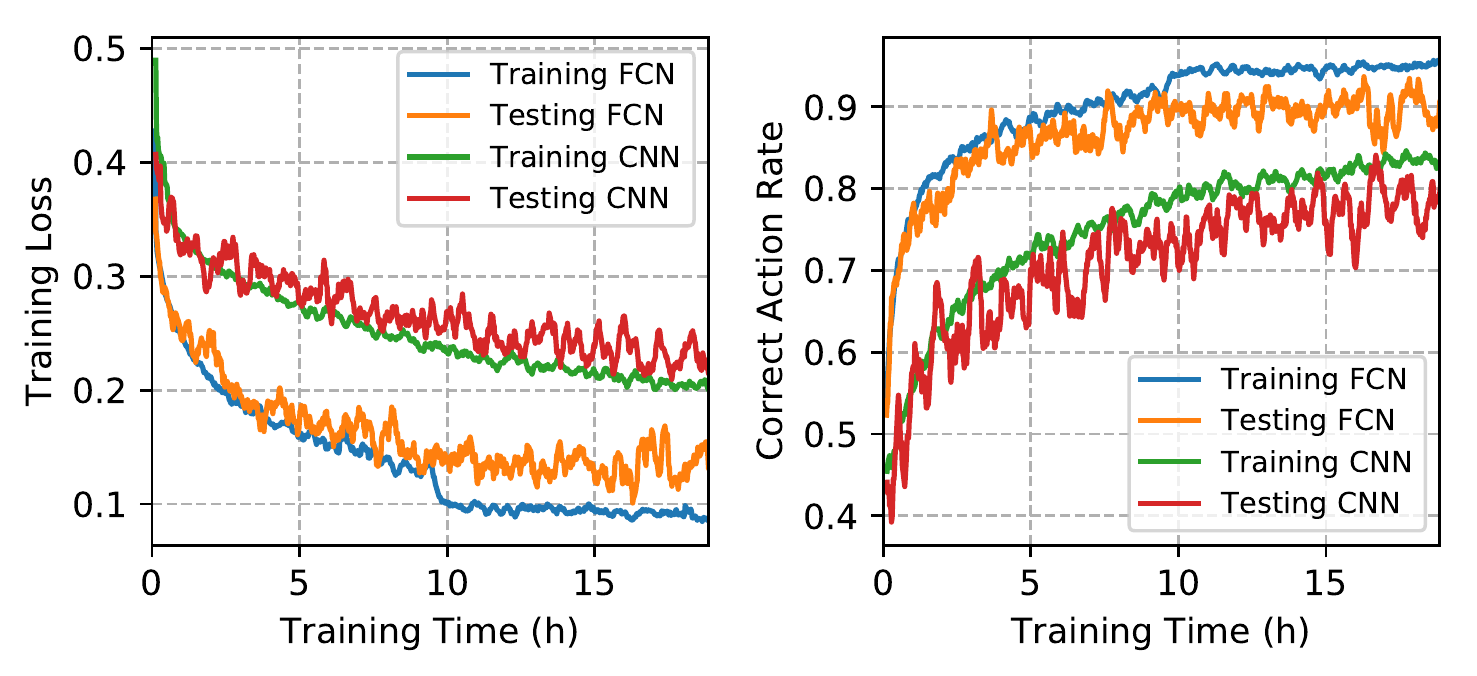}
		\caption{Comparison of training speed using CNN-based training and dilated FCN-based training.}
		\label{fig:train_speed}
	\end{figure}

	We compared training using CNN and the proposed dilated FCN. In the training of CNN, random ROIs are extracted from the X-ray and DRR as the CNN input, and ground truth rewards are calculated and used as supervision. Curves for the training loss and correct action rate using CNN-based and dilated FCN-based training are shown in \refFig{fig:train_speed}. {FCN-based training finished in 17 hours, with a testing loss of $\sim$0.13 and a testing correct action rate of $\sim$90\%. In comparison, CNN-based training after 17 hours reached a test loss of $\sim$0.22 and a testing correct action rate of $\sim$80\%, which is close to the performance of FCN-based method after 2 hours of training.}
	
	\subsection{Testing}
	
	To evaluate the contribution of the multi-agent strategy, we tested the agent-based method in two modes: 1) using a single agent with an ROI at the center of the image, referred to as \textit{agt-s}, and 2) using the proposed multi-agent strategy, referred to as \textit{agt-m}. To apply the proposed method on 2D/3D registration with two X-ray images, in every step, one action is obtained from each X-ray image, and the obtained actions are applied sequentially. We also tested a combination of \textit{agt-m} and an optimization-based method, referred to as \textit{agt-m-opt}, where optimization of GC using BOBYQA optimizer was applied starting from the result of \textit{agt-m}.
	We compared the agent-based method with state-of-the-art optimization-based methods. Multiple similarity measures were evaluated in \cite{de20163d} using CMA-ES optimizer for spine 2D/3D registration, and GC and GO were reported to achieve the best performance. Therefore, we evaluated CMA-ES optimization of GO and GC using the same parameters reported in ~\cite{de20163d}, referred to as \textit{ES-GO} and \textit{ES-GC}, respectively. 
	Registration error is measured by Target Registration Error (TRE), which is calculated as the Root Mean Square Error of the locations of seven anatomical landmarks located on spine vertebrae.
	
	\subsubsection{CBCT Data}
	
	Testing was first performed on 116 CBCT data sets via three-fold cross validation (77 used for training and 39 used for testing). {The typical size of the CBCT data is $512\times 512\times 389$ with a pixel spacing of 0.486 mm}. On each data set, 10 pairs of X-ray images that are $>$60$^\circ$ apart (common practice for spine surgery) were randomly selected, and 2D/3D registration was performed on each pair, starting from a perturbation of the ground truth transformation within 20 mm translation and 10$^\circ$ rotation, leading to 1,160 test cases. Note that X-ray images in CBCT data have a relatively low SNR with a faint spine as shown in \refFig{fig:example_images}. 
	
	Experiment results are summarized in \refTable{tab:rotresult}. The two optimization-based methods, ES-GO and ES-GC, resulted in relatively high gross failure rate (account for TRE$>$10 mm, which is about 1/2 of a vertebrae and considered to be grossly off by our clinical collaborating partners). This is mainly due to the low image quality (e.g., low SNR, image artifacts in CBCT and etc.), which leads to a highly non-convex optimization problem using low level similarity measures like GC and GC. {In comparison, \textit{agt-m} achieved a much lower gross failure rate, demonstrating the robustness advantage of the agent-based method. Comparison of \textit{agt-s} and \textit{agt-m} shows that the multi-agent strategy can noticeably improve robustness by aggregating information from most confident agents.} The comparison of median TRE shows that while the agent-based method provides low failure rate, its accuracy is lower than that of optimization-based methods. This is primarily due to the discrete actions of 1 mm and 1 degree, and location information loss during stride in the CNN. By applying \textit{opt-local} to refine the result of \textit{agt-m}, \textit{apt-m-opt} achieved both low failure rate and high accuracy.
	
	\setlength{\tabcolsep}{4pt}
	\begin{table}[t]
		\centering
		\caption{Experiment results on bi-plane 2D/3D registration on 1,160 test cases from 116 CBCT data sets, and 560 test cases from 28 clinical data sets. Gross failure rate (GFR) accounts for test cases with TRE$>$10 mm. Median, 75th percentile and 95th percentile TREs are reported.}
		\label{tab:rotresult}
		\begin{tabularx}{\linewidth}{llccYYc}
			\toprule
			&
			\multirow{2}[2]{*}{Method} & 
			\multirow{2}[2]{*}{GFR} & 
			\multirow{2}[2]{*}{\makecell{Median \\ (mm)}} & 
			\multicolumn{2}{c}{Percentile (mm)} & 
			\multirow{2}[2]{*}{\makecell{Run \\ Time}} \\ \cmidrule(lr){5-6}
			&&&& 75\% & 95\% & \\\midrule
			\multirow{6}[2]{*}{\rotatebox[origin=c]{90}{CBCT Data}}
			&Start     & 93.4\% & 19.4 & 23.2 & 27.8 & - \\
			\cmidrule{2-7}
			&ES-GC     & 32.2\% & 1.67 & 22.1 & 44.1 & 18.7 s\\
			&ES-GO     & 34.3\% & 1.81 & 21.0 & 38.6 & 33.9 s\\
			\cmidrule{2-7}
			&agt-s     & 17.2\% & 5.30 & 8.13 & 23.3 & 0.5 s \\
			&agt-m     & 4.1\%  & 3.59 & 5.30 & 8.98 & 2.1 s \\
			&agt-m-opt & \textbf{2.1\%} & \textbf{1.19} &\textbf{ 1.76} & \textbf{4.38} & 2.6 s \\
			\midrule
			\multirow{6}[2]{*}{\rotatebox[origin=c]{90}{Clinical Data}}
			&Start     & 95.4\%  & 20.4 & 23.1 & 26.8 & - \\
			\cmidrule{2-7}
			&ES-GC     & 49.3\% & 8.79 & 26.5 & 55.6 & 20.4 s\\
			&ES-GO     & 42.1\% & 3.18 & 29.0 & 84.6 & 35.8 s\\
			\cmidrule{2-7}
			&agt-s     & 45.7\% & 8.42 & 14.2 & 26.4 & 0.6 s \\
			&agt-m     & 6.8\%  & 4.97 & 7.36 & \textbf{10.3} & 2.1 s \\
			&agt-m-opt & \textbf{6.1\%} & \textbf{1.99} & \textbf{2.76} & 10.8 & 2.7 s \\
			\bottomrule
		\end{tabularx}
	\end{table}

	
	\subsubsection{Clinical Data}
	
	To evaluate the proposed method in a real clinical setup, we blindly selected one trained model from the three-hold cross validation on CBCT data, and tested it on 28 clinical data sets collected from minimally invasive spine surgery. Each data set contains a CBCT acquired before the surgery and two X-ray images acquired during the surgery. Ground truth registration was manually annotated by experts. On each data set, 20 perturbations of the ground truth transformation were randomly generated as starting positions for 2D/3D registration, leading to 560 test cases. 
	
	Experiment results on clinical data are summarized in ~\refTable{tab:rotresult}. Higher TREs are reported for all methods on clinical data than that on the CBCT data, primarily due to three reasons: 1) The ground truth registration for clinical data was manually annotated, which could bear 1$\sim$2 mm error; 2) The complexity of clinical data is much higher than the CBCT data (i.e., artifacts and occlusion caused by surgical devices, varying imaging FoVs and etc.); 3) For agent-based methods, the agent was trained without using any real clinical data from spine surgery.
	We observed that due to the increased complexity, the heuristically selected ROI used in \textit{agt-s} (i.e., center of the image) become even less reliable. As a result, the robustness of \textit{agt-s} degrades significantly comparing to that on the CBCT data. The multi-agent system, \textit{agt-m}, in contrast achieved a much higher robustness than \textit{agt-s}, even though the individual agent was trained without using any clinical data from spine surgery, demonstrating the effectiveness of the multi-agent strategy in dealing with complex scenarios.
	
	\section{Conclusion and Future Works}
	
	In this paper, we formulated 2D/3D registration as a MDP, and proposed an multi-agent system to solve this challenging problem. A multi-agent action aggregation scheme is proposed to drive the registration with inherent attention focus. In addition, a dilated FCN-based training scheme is proposed to reduce the number of DoFs that need to be sampled in training from 8 to 4, which speeds up the training efficiency by an order of magnitude. On our experiments on both CBCT data and clinical data, the proposed method is shown to be able to achieve significantly higher robustness than the state-of-the-art 2D/3D registration methods.
	
	In our future works, we will explore the advantages provided by the proposed dilated FCN-based network structure for other general computer vision problems such as optical flow estimation and metric learning. Specifically, unlike FCN-based optical flow (e.g., FlowNet~\cite{dosovitskiy2015flownet}), where the pooling and uppooling layers inherently make the output insensitive to small motion, our network structure ensures that local image descriptor strictly follows the input ROI (i.e., 1 pixel shift of the ROI causes exactly 1 pixel shift of the descriptor), and therefore can potentially provide high accuracy for motion estimation. In addition, this property could be utilized for metric learning using triplet networks, i.e. embeding ROIs to feature vectors with Euclidean distances correlated with the physical distances of the ROIs. The dilated FCN based training scheme could in general be highly efficient by allowing training multiple displacements in each back propagation. These are currently under investigation. 
	
	\bibliographystyle{aaai}
	\bibliography{egbib}
	
\end{document}